\documentclass[runningheads]{llncs}
\usepackage{graphicx}
\usepackage{hyperref}

\usepackage{mathtools}
\usepackage{amssymb}
\usepackage{color}
\usepackage{listings}
\usepackage[caption=false]{subfig}
\usepackage{multirow}
\usepackage{makecell}
\usepackage{caption}
\newcolumntype{L}[1]{>{\raggedright\let\newline\\\arraybackslash\hspace{0pt}}m{#1}}
\newcolumntype{C}[1]{>{\centering\let\newline\\\arraybackslash\hspace{0pt}}m{#1}}
\newcolumntype{R}[1]{>{\raggedleft\let\newline\\\arraybackslash\hspace{0pt}}m{#1}}

\begin{document}
\title{NASTyLinker: NIL-Aware Scalable Transformer-based Entity Linker}
\author{Nicolas Heist\orcidID{0000-0002-4354-9138} \and
Heiko Paulheim\orcidID{0000-0003-4386-8195}}
\authorrunning{N. Heist and H. Paulheim}
\institute{
 Data and Web Science Group, University of Mannheim, Germany
 \email{\{nico,heiko\}@informatik.uni-mannheim.de}
}
\maketitle
\begin{abstract}
Entity Linking (EL) is the task of detecting mentions of entities in text and disambiguating them to a reference knowledge base. Most prevalent EL approaches assume that the reference knowledge base is complete. In practice, however, it is necessary to deal with the case of linking to an entity that is not contained in the knowledge base (NIL entity). Recent works have shown that, instead of focusing only on affinities between mentions and entities, considering inter-mention affinities can be used to represent NIL entities by producing clusters of mentions. At the same time, inter-mention affinities can help to substantially improve linking performance for known entities. With NASTyLinker, we introduce an EL approach that is aware of NIL entities and produces corresponding mention clusters while maintaining high linking performance for known entities. The approach clusters mentions and entities based on dense representations from Transformers and resolves conflicts (if more than one entity is assigned to a cluster) by computing transitive mention-entity affinities. We show the effectiveness and scalability of NASTyLinker on NILK, a dataset that is explicitly constructed to evaluate EL with respect to NIL entities. Further, we apply the presented approach to an actual EL task, namely to knowledge graph population by linking entities in Wikipedia listings, and provide an analysis of the outcome.

\end{abstract}
\section{Introduction}
\subsection{Motivation and Problem}
\begin{figure*}[t]
\subfloat[A lake in Ontario, CA.\\ \textit{Lakes of Temagami}]{%
  \includegraphics[clip,width=.48\textwidth]{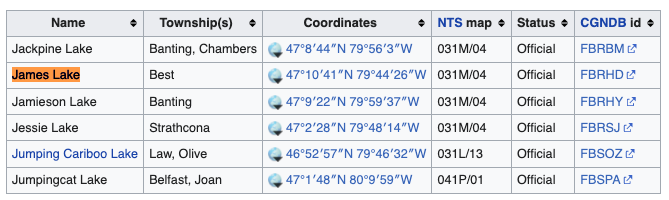}%
  \label{fig:example-mention-1}
}
\hfill
\subfloat[A lake in Montana, US.\\ \textit{List of lakes of Powell County, Montana}]{%
  \includegraphics[clip,width=.48\textwidth]{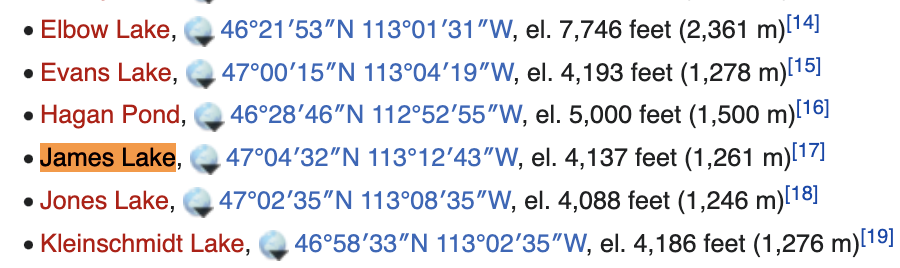}%
  \label{fig:example-mention-2}
}
\hfill
\subfloat[A musician in the band Vinyl Williams.\\ \textit{Vinyl Williams}]{%
  \includegraphics[clip,width=.48\textwidth]{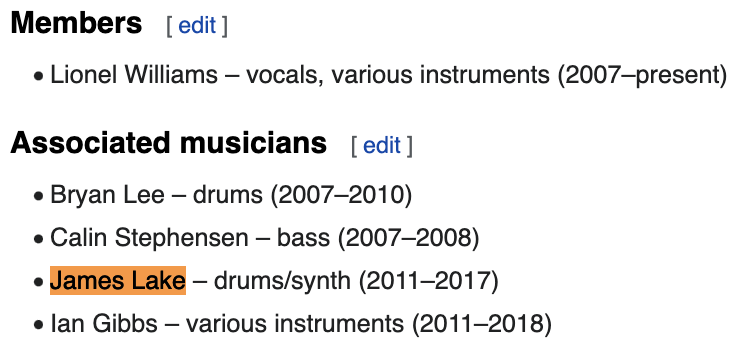}%
  \label{fig:example-mention-3}
}
\hfill
\subfloat[A character in a soap opera.\\ \textit{List of The Young and Restless characters}]{%
  \includegraphics[clip,width=.48\textwidth]{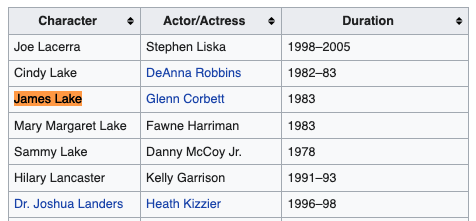}%
  \label{fig:example-mention-4}
}
\caption{Listings in Wikipedia containing the mention \textit{James Lake}. All of the mentions refer to distinct entities. A dedicated Wikipedia page exists only for the entity of the mention in (a).}
\label{fig:running-example}
\end{figure*}

Entity Linking (EL), i.e., the task of detecting mentions of entities in text and disambiguating them to a reference knowledge base (KB), is crucial for many downstream tasks like question answering \cite{das2019multi,wang2021retrieval}, or KB population and completion \cite{heist2020knowledge,heist2021information,paulheim2017knowledge}. One main challenge of EL is the inherent ambiguity of mentioned entities in the text. Figure~\ref{fig:running-example} shows four homonymous mentions of distinct entities with the name \textit{James Lake} (a lake in Canada, a lake in the US, a musician, and a fictional character). Correctly linking the mentions in \ref{fig:example-mention-1} and \ref{fig:example-mention-2} is especially challenging as both point to lakes that are geographically close.

In a typical EL setting, we assume that the training data contains mentions of all entities to be linked against. This assumption is dropped in Zero-Shot EL \cite{logeswaran2019zero}, where a linking decision is made on the basis of entity information in the reference KB (e.g. textual descriptions, types, relations). In this setting, a seminal approach has been introduced with BLINK \cite{wu2020scalable}. Its core idea is to create dense representations of mentions and entities with a Transformer model \cite{devlin2018bert} in a bi-encoder setting, retrieve mention-entity candidates through Nearest Neighbor Search, and rerank candidates with a cross-encoder.

In a practical setting, we additionally encounter the problem of mentions without a corresponding entity in the reference KB (which we refer to as NIL mentions and NIL entities, respectively). In fact, the mention in Figure~\ref{fig:example-mention-1} is the only one with a counterpart in the reference KB (i.e., Wikipedia). For the other mentions, a correct prediction based on Wikipedia entities is impossible. Instead, NIL-aware approaches could either (1) create an (intermediate) entity representation for the NIL entity to link, or (2) produce clusters of NIL mentions with all mentions in a cluster referring to the same entity.

While this problem has been largely ignored by EL approaches for quite some time, recent works demonstrate that reasonable predictions for NIL mentions can be made by clustering mentions on the basis of inter-mention affinities \cite{agarwal2021entity,kassner2022edin}. Both compute inter-mention and mention-entity affinities using a bi-encoder architecture on the basis of BLINK \cite{wu2020scalable}. EDIN \cite{kassner2022edin} is an approach of category (1) that uses a dedicated adaptation dataset to create representations for NIL entities in an unsupervised fashion. Hence, the approach can only link to a NIL entity if there is at least one mention of it in the adaptation dataset. 
For some EL tasks, especially as a prerequisite for KB population, creating an adaptation dataset with good coverage is not trivial because an optimal adaptation dataset has to contain mentions of all NIL entities. Agarwal et al. \cite{agarwal2021entity} present an approach of category (2) that creates clusters of mentions and entities in a bottom-up fashion by iteratively merging the two most similar clusters, always under the constraint that a cluster must contain at most one entity.

\subsection{Approach and Contributions}
With NASTyLinker, we present an EL approach that is NIL-aware in the sense of category (2) and hence avoiding the need for an adaptation dataset. Similar to Agarwal et al. \cite{agarwal2021entity}, it produces clusters of mentions and entities on the basis of inter-mention and mention-entity affinities from a bi-encoder. NASTyLinker relies on a top-down clustering approach that -- in case of a conflict -- assigns mentions to the entity with the highest transitive affinity. Contrary to Agarwal et al., who discard cross-encoders completely due to the quadratic growth in complexity when evaluating inter-mention affinities, our experiments show that applying a cross-encoder only for the refinement of mention-entity affinities can result in a considerable increase of linking performance at a reasonable computational cost. Our evaluation on the NILK dataset \cite{iurshina2022nilk}, a dataset especially suited for the evaluation of NIL-aware approaches, shows that NASTyLinker manages to make competitive predictions for NIL entities while even slightly improving prediction performance for known entities. The approach is designed in a modular way to make existing EL models NIL-aware by post-processing the computed inter-mention and mention-entity affinities. By applying NASTyLinker to a knowledge graph population task, we demonstrate its ability to reliably link to known entities (up to 87\% accuracy) and identify NIL entities (up to 90\% accuracy). 

To summarize, the contributions of this paper are as follows:
\begin{itemize}
    \item We introduce the NASTyLinker approach, serving as an extension to existing EL approaches by using a top-down clustering mechanism to consistently link mentions to known entities and produce clusters for NIL mentions (Section~\ref{sec:nastylinker}).
    \item In our experiments, we demonstrate the competitive linking performance and scalability of the presented approach through an evaluation on the NILK dataset (Section~\ref{sec:entity-linking-performance}).
    \item We use NASTyLinker for KB population by linking entities in Wikipedia listings. We report on the linking statistics and provide a qualitative analysis of the results (Section~\ref{sec:linking-entities-in-wikipedia-listings}).
\end{itemize}

The produced code is part of the CaLiGraph extraction framework and publicly available on GitHub.\footnote{\url{https://github.com/nheist/CaLiGraph}}
\section{Related Work}
\textbf{Entity Linking }
Entity Linking has been studied extensively in the last two decades \cite{rao2013entity,sevgili2022neural}. Initially, approaches relied on word and entity frequencies, alias tables, or neural networks for their linking decisions \cite{cucerzan2007large,ganea2017deep,milne2008learning}. The introduction of pre-trained transformer models \cite{devlin2018bert} made it possible to create representations of mentions and entities from text without relying on other intermediate representations. Gillick et al. \cite{gillick2019learning} show how to learn dense representations for mentions and entities, Logeswaran et al. \cite{logeswaran2019zero} extend this by introducing the zero-shot EL task and demonstrating that reasonable entity embeddings can be derived solely from entity descriptions. Wu et al. \cite{wu2020scalable} introduce BLINK, the prevalent bi-encoder and cross-encoder paradigm for zero-shot EL. Various improvements for zero-shot EL have been proposed based on this paradigm. KG-ZESHEL \cite{ristoski2021kg} adds auxiliary entity information from knowledge graph embeddings into the linking process; Partalidou et al. \cite{partalidou2022improving} propose alternative pooling functions for the bi-encoder to increase the accuracy of the candidate generation step.

\textbf{Cross-Document Coreference Resolution }
NIL-Aware EL is closely related to Cross-Document Coreference Resolution (CDC), the task of identifying coreferent entity mentions in documents without explicitly linking them to entities in a KB \cite{bagga1998entity}. Dutta and Weikum \cite{dutta2015c3el} explicitly tackle CDC in combination with EL by applying clustering to bag-of-words representations of entity mentions. More recently, Logan IV et al. \cite{logan2021benchmarking} evaluate greedy nearest-neighbour and hierarchical clustering strategies for CDC, however, without explicitly evaluating them with respect to EL.

\textbf{Entity Discovery and NIL-Aware EL }
The majority of EL approaches may identify NIL mentions (for instance, through a binary classifier or a ranking that explicitly includes \textit{NIL}), but does not process them in any way \cite{sevgili2022neural,shen2014entity}. In 2011, the TAC-KBP challenge \cite{ji2010overview} introduced a task that includes NIL clustering; in the NEEL challenge \cite{rizzo2017lessons} that is based on microposts, NIL clustering was part of the task as well. Approaches that tackled these tasks typically applied clustering based on similarity measures over the entity mentions in the text \cite{blissett2019cross,fahrni2013hits,greenfield2016reverse,monahan2011cross,radford2011naive}. More recently, Angell et al. \cite{angell2021clustering} train two separate bi-encoders and cross-encoders to compute inter-mention and mention-entity affinities. Subsequently, they apply a bottom-up clustering for refined linking predictions within single biomedical documents. Agarwal et al. \cite{agarwal2021entity} extend the approach to cross-document linking through a clustering based on minimum spanning trees over all mentions in the corpus. Clusters are formed by successively adding edges to a graph as long as the constraint that a cluster can contain at most one entity is not violated. They omit the cross-encoder and employ a custom training procedure for the bi-encoder instead. They explicitly evaluate their approach w.r.t. NIL entity discovery by removing a part of the entities in the training set from zero-shot EL benchmark datasets. In our approach, we employ a similar method for computing affinities but employ a top-down clustering approach that aims to better identify clusters of NIL mentions. The EDIN pipeline \cite{kassner2022edin} also applies clustering w.r.t. inter-mention and mention-entity affinities, but only to identify NIL mention clusters on a dedicated adaptation dataset. Subsequently, the entity index is enhanced with pooled representations of these clusters to make a prediction of NIL entities possible. In their clustering phase, they first produce groups of mentions and then identify NIL mention clusters by checking whether less than 70\% of the mentions are referring to the same entity. As we aim to apply NIL-aware EL for KB population, relying on an adaptation dataset is not possible. Still, we include the clustering method of the EDIN pipeline in our experiments to compare how well the approaches detect NIL mention clusters.
\section{Task Formulation}
A document corpus $\mathcal{D}$ contains a set of textual entity mentions $\mathcal{M}$. Each of the mentions $m \in \mathcal{M}$ refers to an entity $e$ in the set of all entities $\mathcal{E}$. Given a knowledge base K with known entities $\mathcal{E}^k$, the task in standard EL is to assign an entity $\hat{e} \in \mathcal{E}^k$ to every mention in $\mathcal{M}$. In this setting, we assume that $\mathcal{E}=\mathcal{E}^k$, i.e., all entities are contained in K. 

In NIL-aware EL, we drop the assumption that every mention links to an entity contained in K. Instead there is a set of NIL entities $\mathcal{E}^n$ with $\mathcal{E}^k \cup \mathcal{E}^n = \mathcal{E}$ and $\mathcal{E}^k \cap \mathcal{E}^n = \emptyset$. For mentions $\mathcal{M}^k$ that refer to entities in K, the task is still to predict an entity $\hat{e} \in \mathcal{E}^k$. For mentions $\mathcal{M}^n$ that refer to entities not contained in K, the task is to predict a cluster identifier $c \in \mathcal{C}$ so that the clustering $\mathcal{C}$ resembles the distribution of mentions in $\mathcal{M}^n$ to entities in $\mathcal{E}^n$ as closely as possible. We assume that we are additionally operating in a zero-shot setting, i.e., the training portion $\mathcal{D}_{train}$ of the document corpus may not contain mentions for all entities in $\mathcal{E}$.

Note that, similar to related works \cite{agarwal2021entity,logeswaran2019zero}, we assume that the textual entity mentions are already given. Further, we only investigate the relevant steps for KB population, i.e., detection and disambiguation of NIL entities. While we discard the indexing aspect, an EL model which includes the entities in $\mathcal{E}^n$ can still be created in a subsequent step by training a new model on the enhanced KB.
\section{NASTyLinker: An Approach for NIL-Aware and Scalable Entity Linking}
\label{sec:nastylinker}
\begin{figure}[t]
  \centering
  \includegraphics[width=\linewidth]{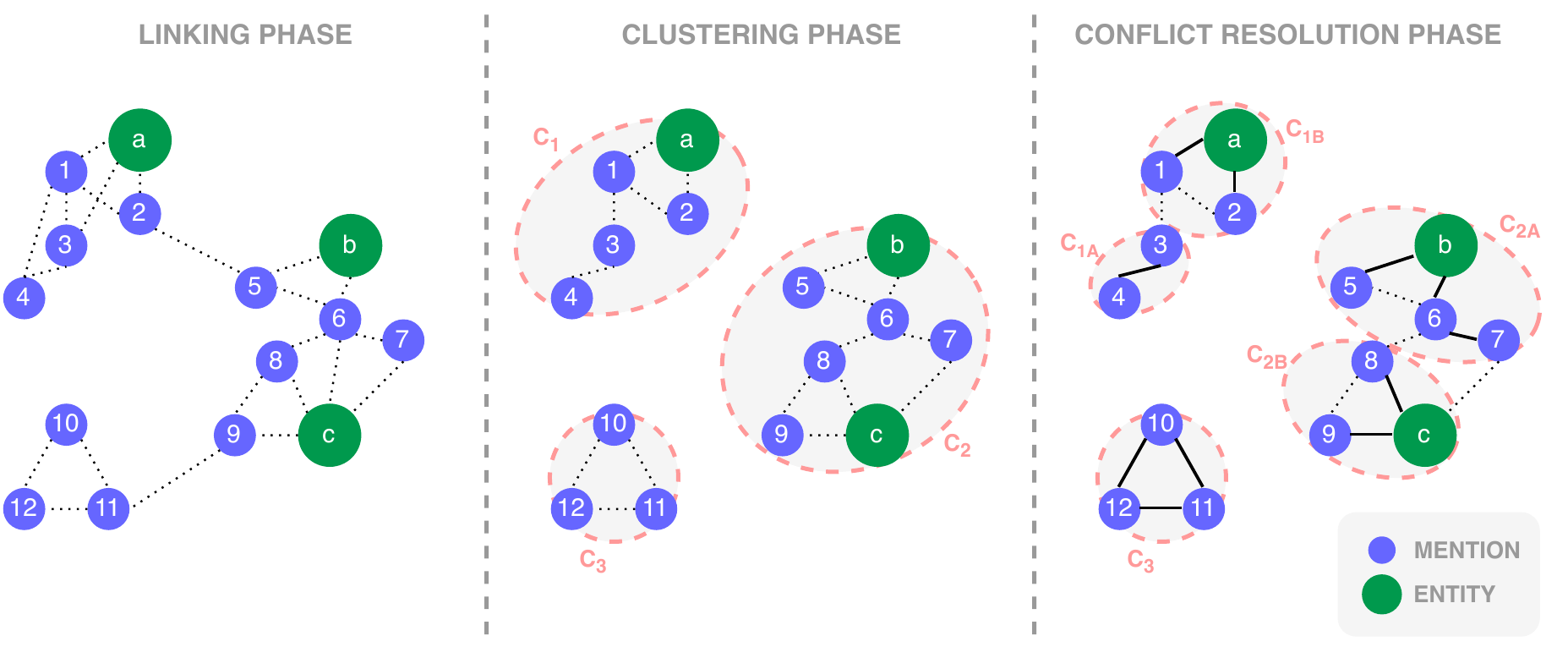}
  \caption{Main phases of the NASTyLinker approach. Dotted lines show top-k affinity scores, solid lines indicate the highest transitive affinity scores.}
  \label{fig:approach-overview}
\end{figure}

In this section, we describe our proposed approach for making NIL-aware EL predictions. Figure~\ref{fig:approach-overview} depicts the three main phases of the NASTyLinker approach. In the \textit{Linking Phase}, we first retrieve inter-mention and mention-entity affinities from an underlying EL model for the subsequent clustering. We define constraints for such a model and describe the one used in our experiments in Section~\ref{sec:entity-linking-model}. During the \textit{Clustering Phase}, clusters of mentions and entity candidates are created using greedy nearest-neighbour clustering (Section~\ref{sec:cluster-initialization}). Finally, we retrieve entity candidates for every cluster. In the \textit{Conflict Resolution Phase}, clusters are split based on transitive mention-entity affinities to ensure that a cluster contains at most one known entity (Section~\ref{sec:cluster-conflict-resolution}).
\subsection{Entity Linking Model}
\label{sec:entity-linking-model}
In the \textit{Linking Phase} we compute the \textit{k} most similar mentions and entities for every mention in $\mathcal{M}$ (dotted lines in Figure~\ref{fig:approach-overview}). The underlying EL model has to provide a function $\phi$ with $\phi(m,e) \in [0;1]$ for the similarity between mention $m$ and entity $e$ as well as $\phi(m,m') \in [0;1]$ for the similarity between mentions $m$ and $m'$. In addition to that, it must be possible to retrieve the top \textit{k} mention and entity candidates for a given mention in an efficient manner.

For our experiments with NASTyLinker, we choose the BLINK architecture \cite{wu2020scalable} as the underlying EL model as it provides the foundation for many state-of-the-art EL models. Furthermore, as the bi-encoder creates embeddings for mentions and entities alike, methods for an approximate nearest neighbour search like FAISS \cite{johnson2019billion} can be used to retrieve linking candidates efficiently. As the application of the cross-encoder is the most time-consuming part of this model, we explore in our experiments the trade-off between linking performance and runtime when reranking only inter-mention affinities, only mention-entity affinities, or both.

Partalidou et al. \cite{partalidou2022improving} propose several layouts for structuring the input sequence of mentions and entities for the Transformer model. We achieved the best results with the mention layout
\begin{verbatim}
    [CLS] [<type>] <mention label> [CTX] <mention context> [SEP]
\end{verbatim}
and the entity layout
\begin{verbatim}
    [CLS] [<type>] <entity title> [CTX] <entity description> [SEP]
\end{verbatim}
where \texttt{[CTX]} is a special delimiter token and \texttt{[<type>]} is a placeholder for a special token of the mention type (POS-tag) or entity type (top-level type in the KB). For optimization, we stick to Wu et al. \cite{wu2020scalable} and use in-batch (hard) negatives for the bi-encoder, and bi-encoder-generated negatives for the cross-encoder.
\subsection{Cluster Initialization}
\label{sec:cluster-initialization}
To produce an initial mention clustering, we follow Logan IV et al. \cite{logan2021benchmarking} and use a greedy nearest-neighbour clustering. Given the mention affinity threshold $\tau_m$, the mentions $\mathcal{M}$ are grouped into clusters $\mathcal{C}$ so that two mentions $m, m' \in \mathcal{M}$ belong to the same cluster if $\phi(m,m') > \tau_m$.

Further, we assign entity candidates to the clusters using a threshold for entity affinity $\tau_e$. For a cluster $C \in \mathcal{C}$ with mentions $M_c$, we select the known entities with the highest affinity to each cluster mention:
\begin{equation}
    E^k_c = \bigcup_{m \in M_c}\{\mathop{argmax}\limits_{e \in \mathcal{E}^k} \phi(m,e) : \phi(m,e) > \tau_e\}.
\end{equation}
In Figure~\ref{fig:approach-overview}, the dotted lines represent affinities greater than the thresholds $\tau_m$ and $\tau_e$, respectively. Cluster $C_1$ contains four loosely connected mentions with $m_1$ and $m_2$ directly connected to the entity candidate $e_a$. Either all four mentions refer to $e_a$ as they are transitively connected, or some mentions refer to an entity in $\mathcal{E}^n$ (e.g., a situation like in Figure~\ref{fig:example-mention-1} and \ref{fig:example-mention-2}). Cluster $C_2$ contains several mentions with two known entity candidates $e_b$ and $e_c$, making a trivial assignment of mentions to entities impossible. Finally, cluster $C_3$ contains three connected mentions without any assigned entity candidates, most likely representing a NIL entity. Conflicts like the ones occurring in the former two clusters are resolved in the subsequent resolution phase.
\subsection{Cluster Conflict Resolution}
\label{sec:cluster-conflict-resolution}
The objectives of the \textit{Conflict Resolution Phase} are twofold: For every cluster $C \in \mathcal{C}$ we (1) find sub-clusters with $|E^k_c| = 1$ (c.f. $C_{1B}$, $C_{2A}$, and $C_{2B}$ in Figure~\ref{fig:approach-overview}), and (2) identify mentions in $M_c$ that do not refer to any entity in $E^k_c$. For these, we create one or more sub-clusters representing the NIL entities $E^n_c$ of $C$ (c.f. $C_{1A}$ and $C_3$ in Figure~\ref{fig:approach-overview}).

For conflict resolution, we view a cluster $C \in \mathcal{C}$ as a graph $G_c$ with $M_c \cup E^k_c$ as nodes, and affinities above threshold as edges. To ensure objective (1), we assign every mention in a cluster to the candidate entity with the highest transitive, defined as follows:
\begin{equation}
    \phi^*(m,e) = \mathop{max}\limits_{m \sim e \in G_c} \prod_{u,v}^{m \sim e} \phi(u,v)
\end{equation}
with $m \sim e$ denoting a path from a mention $m$ to an entity $e$ in $G_c$ and $(u,v)$ a single edge. The rationale for this metric is to favour strong contextual similarity between mentions over the mediocre similarity between a mention and an entity. As the entity context is coming from a different data corpus (i.e., information from a KB) than the mention context, it is more likely to happen that the contexts for a mention and its linked entity are dissimilar than the contexts of two mentions linking to the same entity.

\textbf{Example 1} With affinities $\phi(m_6,e_b) = 0.9$, $\phi(m_6,m_7) = 0.9$, $\phi(m_7,e_c) = 0.8$, and paths $m_7-m_6-e_b$, $m_7-e_c$ from Figure~\ref{fig:approach-overview}, we find that $\phi^*(m_7,e_b) = 0.81 > \phi^*(m_7,e_c) = 0.8$, resulting in the assignment of $m_7$ to the cluster of $e_b$ in spite of $e_c$ being the most likely entity for $m_7$ w.r.t. $\phi$.

To ensure objective (2), we introduce a threshold $\tau_a$ as a lower limit for the transitive affinity between a mention and an entity. We label mentions as NIL mentions if they do not have a transitive affinity higher than the threshold to any entity in $E^k_c$:
\begin{equation}
M^n_c = \{m \in M_c | \nexists e \in E^k_c: \phi^*(m,e) > \tau_a \}
\end{equation}
From $M^n_c$ we produce one or more mention clusters similar to the initialization step in Section~\ref{sec:cluster-initialization}.

\textbf{Example 2} With $\tau_a = 0.75$, affinities $\phi(m_1,e_a) = 0.9$, $\phi(m_1,m_3) = 0.8$, $\phi(m_3,m_4) = 0.9$, and path $m_4-m_3-m_1-e_a$ from Figure~\ref{fig:approach-overview}, we find that $\phi^*(m_3,e_a) = 0.72 < \tau_a$ and $\phi^*(m_4,e_a) = 0.648 < \tau_a$. $m_3$ and $m_4$ are labelled as NIL mentions and form - due to their direct connection - the single cluster $C_{1B}$.

The function $\phi^*$ can be computed efficiently on a graph using Dijkstra's algorithm with $-log\phi$ as a function for edge weights. Edges are only inserted in the graph for $\phi>\tau_a$, avoiding undefined edge weights in the case of $\phi=0$.
\section{Experiments}
\label{sec:experiments}
We first describe the datasets and experimental setup used for the evaluation of NASTyLinker. Then, we compare the performance of our approach with related NIL-aware clustering approaches on the NILK dataset \cite{iurshina2022nilk} and analyze its potential to scale. Finally, we report on the application of NASTyLinker for KB population by linking entities in Wikipedia listings.
\subsection{Datasets}
\label{sec:datasets}
{\renewcommand{\arraystretch}{1.25}%
    \begin{table}
      \centering
      \caption{Mention and entity occurrences in the partitions of the datasets. NIL mention counts for $\mathcal{D}^L$ are estimated w.r.t. partial completeness assumption. Furthermore, the number of NIL entities $\mathcal{E}^n$ in the listings dataset is not known. For $\mathcal{D}^L_{pred}$ a single mention count is displayed as we cannot know whether a mention in $\mathcal{M}$ links to an entity in $\mathcal{E}^k$ or $\mathcal{E}^n$.}
      \label{tab:datasets}
      \begin{tabular}{|C{0.6cm}|L{2.7cm}|C{2.05cm}|C{2.05cm}|C{2.05cm}|C{2.05cm}|}
        \hline
        \multicolumn{2}{|c|}{\textbf{Dataset}} & \textbf{$|\mathcal{M}^k|$} & \textbf{$|\mathcal{M}^n|$} & \textbf{$|\mathcal{E}^k|$} & \textbf{$|\mathcal{E}^n|$}\\
        \hline \hline
        \parbox[t]{2mm}{\multirow{3}{*}{\rotatebox[origin=c]{90}{\textbf{NILK}}}} & Training ($\mathcal{D}^{N}_{train}$) & 85,052,764 & \hphantom{1}1,327,039 & 3,382,497 & 282,210 \\
        & Validation ($\mathcal{D}^{N}_{val}$) & 10,525,107 & \hphantom{11,}162,948 & \hphantom{1,}422,812 & \hphantom{1}35,276 \\
        & Test ($\mathcal{D}^{N}_{test}$) & 10,451,126 & \hphantom{11,}162,497 & \hphantom{1,}422,815 & \hphantom{1}35,279 \\
        \hline \hline
        \parbox[t]{2mm}{\multirow{4}{*}{\rotatebox[origin=c]{90}{\textbf{LISTING}}}} & Training ($\mathcal{D}^{L}_{train}$) & 11,690,019 & \hphantom{1}6,760,273 & 3,073,238 & ? \\
        & Validation ($\mathcal{D}^{L}_{val}$) & \hphantom{1}3,882,641 & \hphantom{1}2,272,941 & 1,695,156 & ? \\
        & Test ($\mathcal{D}^{L}_{test}$) & \hphantom{1}3,884,066 & \hphantom{1}2,259,072 & 1,701,015 & ? \\
        & Prediction ($\mathcal{D}^{L}_{pred}$) & \multicolumn{2}{|c|}{18,658,271} & ? & ? \\
        \hline
    \end{tabular}
    \end{table}
}
\subsubsection{NILK}
NILK is a dataset that is explicitly created to evaluate EL both for known and NIL entities. It uses Wikipedia as a text corpus and Wikidata \cite{vrandevcic2014wikidata} as reference KB. All entities contained in Wikidata up to 2017 are labelled as known entities and entities added to Wikidata between 2017 and 2021 are labelled as NIL entities. Mention and entity counts of NILK are displayed in Table~\ref{tab:datasets}. About 1\% of mentions in NILK are NIL mentions, and about 6\% of entities are NIL entities. NIL entities are probably slightly biased towards more popular entities, as the fact that they are present in Wikidata hints at a certain popularity, which may be higher than the popularity of an average NIL entity. Hence, the average number of mentions per NIL entity is quite high in this dataset: half of the entities are mentioned more than once, and more than 15\% are even mentioned more than 5 times. Mention boundaries are already given and the authors define partitions for training, validation, and test, which are split in a zero-shot manner w.r.t. NIL entities. As mention context, the authors provide 500 characters before and after the actual mention occurrence in a Wikipedia page. As entity descriptions, we use Wikipedia abstracts.\footnote{While there are entities in Wikidata which do not have a Wikipedia page, this case does not occur in NILK by construction.}
\subsubsection{Wikipedia Listings}
The LISTING dataset was extracted in prior work \cite{heist2022transformer} and consists of entity mentions in enumerations and tables of Wikipedia. Instead of all possible mentions, the focus is only on \textit{subject entities}, which we define as \textit{all entities in a listing appearing as instances to a common concept} \cite{heist2021information}. So every item in a listing is assumed to have one main entity the item is about. For example, in Fig.~\ref{fig:example-mention-4}, the soap opera characters are considered entity mentions, while the actors are not.

As reference KB we use CaLiGraph \cite{heist2019uncovering,heist2020entity}. Mention and entity statistics are given in Table~\ref{tab:datasets}. We partition the data into train, validation, and test while making sure that listings on a page are all in the same split. Contrary to NILK, the LISTING dataset does not contain explicit labels for NIL entities. Instead, we define NIL entities using the partial completeness assumption (PCA). Given a listing with multiple mentions, we only incorporate them into training or test data if at least one mention is linked to a known entity. Then, by PCA, we assume that all mentions that can be linked are actually linked. All other mentions are assigned a new unique entity identifier. The prediction partition $\mathcal{D}^L_{pred}$, however, contains all mentions without a linked entity (i.e., they may link to a known or to a NIL entity). We use the text of the listing item as mention context for the dataset, and we use Wikipedia abstracts as entity descriptions.

We have considered further datasets that were used for evaluation of NIL-aware approaches for evaluation (e.g. from challenges like TAC-KBP or Microposts \cite{DBLP:conf/msm/2016}), but discarded them due to their small size or not being free to use.
\subsection{Metrics}
\subsubsection{Classification Metrics}
We compute precision, recall, and F1-score as well as aggregations of the metrics on the instance level (micro average). As the evaluated approaches are not aware of the true NIL entities, they assign cluster identifiers to (what they assume to be) NIL mentions. To compute the classification metrics, it is necessary to map the cluster identifiers to actual NIL entities. Kassner et al. \cite{kassner2022edin} allow the assignment of multiple cluster identifiers to the same NIL entity. This assumption would yield overly optimistic results. Instead, we only allow one-to-one mappings between cluster identifiers and NIL entities. Finding an optimal assignment for this scenario is equivalent to solving the linear sum assignment problem \cite{alfaro2022assignment}, for which efficient algorithms exist.
\subsubsection{Clustering Metrics}
Following related approaches \cite{agarwal2021entity,kassner2022edin}, we additionally provide normalized mutual information (NMI) and adjusted rand index (ARI) as clustering metrics for the comparison of the approaches to settings where no gold labels of NIL entities may be available.\footnote{We implement further clustering metrics (B-Cubed+, CEAF, MUC) but do not list them as they are similar to or adaptations of the classification metrics.} For known entities, however, the classification metrics will most likely be more expressive than the clustering metrics as the latter treat multiple clusters with the same known entity as their label still as separate clusters.
\subsection{Evaluated Approaches}
\subsubsection{EL Model}
We compute inter-mention and mention-entity affinities with a bi-encoder similar to BLINK \cite{wu2020scalable}. As the reranking of bi-encoder results with a cross-encoder is costly, we evaluate different scenarios where the cross-encoder is omitted (\textit{No Reranking}), applied to inter-mention affinities only (\textit{Mention Reranking}), applied to mention-entity affinities only (\textit{Entity Reranking}), or applied to both (\textit{Full Reranking}). We use the Sentence-BERT implementation of the bi-encoder and cross-encoder \cite{reimers2019sentence} with \textit{all-MiniLM-L12-v2} and \textit{distilbert-base-cased} as respective base models. The base models are fine-tuned for at most one million steps on the training partitions of the datasets. Longer fine-tuning did not yield substantial improvements. We use a batch size of 256 for the bi-encoder and 128 for the cross-encoder. For efficient retrieval of candidates from the bi-encoder, we apply approximate nearest neighbour search with hnswlib \cite{malkov2018efficient}.

We use the plain bi-encoder and cross-encoder predictions of the EL model as baselines. Additionally, we evaluate a trivial \textit{Exact Match} approach, where we link a mention to an entity if their textual representations match exactly.\footnote{We apply simple preprocessing like lower-casing and removal of special characters.} In case of multiple matches, the more popular entity (w.r.t. ingoing and outgoing links in the KB) is selected. Naturally, this approach cannot handle NIL entities.
\subsubsection{Clustering Approaches}
Apart from the NASTyLinker clustering as described in Section~\ref{sec:nastylinker}, we apply the clustering approaches of Kassner et al. \cite{kassner2022edin} and Agarwal et al. \cite{agarwal2021entity} for comparison.\footnote{We tried to compare with the full approach of Agarwal et al. but they do not provide any code and our efforts to re-implement it did not yield improved results.} The clustering approach of Kassner et al., which we call \textit{Majority Clustering}, applies a greedy clustering and assigns a known entity $e$ to a cluster if at least 70\% of mentions in the cluster have the highest affinity to $e$. Similarly to NASTyLinker, they use hyperparameters as thresholds for minimum inter-mention and mention-entity affinities.

The clustering approach of Agarwal et al., which we call \textit{Bottom-Up Clustering}, starts with an empty graph and iteratively adds the edge with the highest affinity, as long as it does not violate the constraint of a cluster having at most one entity. They use a single hyperparameter as a threshold for the minimum affinity of an edge, be it inter-mention or mention-entity.
\subsubsection{Hyperparameter Tuning}
We select the hyperparameters of the EL model ($k, learning\_rate, warmup\_steps$) and the thresholds of all three clustering approaches w.r.t. micro F1-score on the validation partition of the datasets. For a fair comparison, we also test multiple values for the threshold for entity assignment of Majority Clustering, which in the original paper was fixed at 0.7.

Our experiments are run on a single machine having 96 CPUs, 1TB of RAM, and an NVIDIA RTX A6000 GPU with 48GB of RAM.
\subsection{Entity Linking Performance}
\label{sec:entity-linking-performance}
We tune hyperparameters by evaluating on $\mathcal{D}^N_{val}$. For the EL model, we use a \textit{k} of 4, a learning rate of 2e-5, and no warmup steps. For $\tau_m$, a value between 0.8 and 0.9 works best for all approaches. For $\tau_e$, the best values revolve around 0.9 for NASTyLinker and Bottom-Up Clustering, and around 0.8 for Majority Clustering. We use an affinity threshold $\tau_a$ of 0.75 for NASTyLinker and find that the 0.7 threshold of Majority Clustering produces the best results.

{\renewcommand{\arraystretch}{1.05}%
    \begin{table}
      \centering
      \caption{Results for the test partition $\mathcal{D}^{N}_{test}$ of the NILK dataset.}
      \label{tab:nilk-results}
      \begin{tabular}{|C{2cm}|C{2.5cm}|C{0.7cm}|C{0.7cm}|C{0.7cm}|C{0.7cm}|C{0.7cm}|C{0.7cm}|C{0.7cm}|C{0.7cm}|C{0.7cm}|}
        \hline
        \multicolumn{2}{|c|}{\textbf{Approach}} & \multicolumn{3}{c|}{\textbf{Known}} & \multicolumn{3}{c|}{\textbf{NIL}} & \multicolumn{3}{c|}{\textbf{Micro}} \\
        \multicolumn{2}{|c|}{} & F1 & NMI & ARI & F1 & NMI & ARI & F1 & NMI & ARI \\
        \hline \hline
        \multirow{3}{*}{\shortstack{\textbf{No}\\\textbf{Clustering}}} & Exact Match & 79.5 & --- & --- & \hphantom{0}0.0 & --- & --- & 78.1 & --- & --- \\
         & Bi-Encoder & 80.8 & --- & --- & \hphantom{0}0.0 & --- & --- & 79.1 & --- & --- \\
         & Cross-Encoder & 89.0 & --- & --- & \hphantom{0}0.0 & --- & --- & 87.1 & --- & --- \\
        \hline
        \multirow{3}{*}{\shortstack{\textbf{Clustering}\\\textbf{\& No}\\\textbf{Reranking}}}
         & Bottom-Up & 64.6 & 99.0 & 97.5 & 41.6 & 94.8 & 81.8 & 64.1 & 96.8 & 93.5 \\
         & Majority & 59.4 & 99.3 & 98.0 & 49.8 & 92.7 & 82.7 & 59.2 & 96.6 & \textbf{94.6} \\
         & NASTyLinker & 76.8 & 98.6 & 95.3 & 40.8 & \textbf{95.2} & 76.8 & 76.0 & 97.3 & 90.3 \\
        \hline
        \multirow{3}{*}{\shortstack{\textbf{Clustering}\\\textbf{\& Mention}\\\textbf{Reranking}}}
         & Bottom-Up & 65.7 & 97.1 & 98.9 & 41.5 & 94.6 & 10.0 & 65.1 & 96.0 & 66.0 \\
         & Majority & 66.6 & 92.4 & 74.8 & 44.0 & 94.4 & 73.2 & 66.1 & 92.4 & 70.4 \\
         & NASTyLinker & 74.2 & 99.0 & 96.6 & 39.2 & 85.6 & 16.5 & 73.5 & 95.5 & 81.6 \\
        \hline
        \multirow{3}{*}{\shortstack{\textbf{Clustering}\\\textbf{\& Entity}\\\textbf{Reranking}}}
         & Bottom-Up & 89.0 & 99.3 & 96.2 & 41.6 & 94.1 & 58.0 & 87.9 & 98.2 & 92.6 \\
         & Majority & 74.2 & 99.1 & \textbf{99.3} & \textbf{54.1} & 89.3 & \textbf{92.5} & 73.7 & 96.6 & \textbf{94.6} \\
         & NASTyLinker & \textbf{90.4} & 99.3 & 95.5 & 43.7 & 94.6 & 85.3 & \textbf{89.4} & \textbf{98.5} & 84.1 \\
         \hline
        \multirow{3}{*}{\shortstack{\textbf{Clustering}\\\textbf{\& Full}\\\textbf{Reranking}}}
         & Bottom-Up & 84.2 & \textbf{99.6} & 98.9 & 41.8 & 84.6 & 3.2 & 83.3 & 96.2 & 65.5 \\
         & Majority & 80.3 & 95.1 & 95.9 & 51.7 & 90.0 & 39.2 & 79.6 & 93.9 & 70.4 \\
         & NASTyLinker & 87.9 & 99.5 & 99.2 & 42.5 & 87.6 & 33.6 & 86.9 & 97.4 & 71.7 \\
        \hline
    \end{tabular}
    \end{table}
}

\subsubsection{NILK Results}
As shown in Table~\ref{tab:nilk-results}, we evaluate all clustering approaches on $\mathcal{D}^N_{test}$ in different reranking scenarios. We find Exact Match already to be a strong baseline for known entities with an F1 of 79.5\%, which the Cross-Encoder outperforms by approximately 10\%. Even without reranking, the three clustering approaches are able to achieve an F1-score between 40\% and 50\% for NIL entities. Overall, Majority Clustering is best suited to identify NIL entities. It is the only one to substantially benefit from reranking, increasing the F1-score by 10\% when applying entity reranking. Especially for linking known entities, applying only entity reranking is the most favourable scenario, leading even to slight improvements over the baseline approaches that focus only on known entities. 

As the reranking of mentions tends to lead to a decrease in results while considerably increasing runtime, we omit mention reranking (and hence, full reranking) in experiments with Wikipedia listings. In the remaining scenarios, NASTyLinker finds the best balance between the linking of known entities and the identification of NIL entities w.r.t. F1-score and NMI.
\subsubsection{Runtime and Scalability}
The fine-tuning of the bi-encoder and cross-encoder models took 2 hours and 12 hours, respectively. For prediction with a k of 4 on $\mathcal{D}^N_{test}$, the bi-encoder needed 6 hours. Reranking entity affinities with the cross-encoder took 38 hours. Clustering the results with any of the three approaches took an additional 8 to 12 minutes.

\begin{figure}[t]
  \centering
  \caption{Runtime of NASTyLinker components for predictions on samples of $\mathcal{D}^N_{test}$.}
  \includegraphics[width=\linewidth]{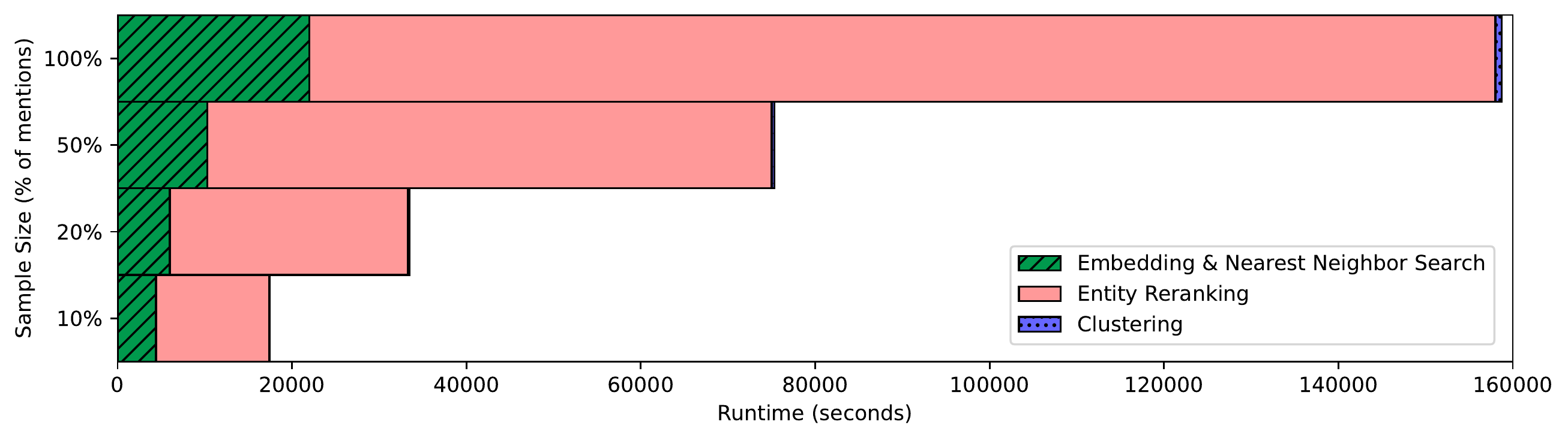}
  \label{fig:scalability}
\end{figure}

In Figure~\ref{fig:scalability} we give an overview of the runtime of NASTyLinker components, compared over various sample sizes of $\mathcal{D}^N_{test}$. Overall, we can see that the total runtime scales linearly.
With a smaller sample size, the computation of embeddings and nearest neighbour search with the bi-encoder is responsible for a larger fraction of the total runtime. We find that this is due to the relatively large overhead of creating the index for the approximate nearest neighbour search. With increasing sample size, this factor is less important for the overall runtime. In general, entity reranking is responsible for most of the total runtime.

The runtime of the clustering itself is responsible for approximately 1\% of total runtime and we do not expect it to increase substantially, as Dijkstra's algorithm has log-linear complexity and the size of mention clusters can be controlled by the threshold $\tau_m$. Hence, the runtime of NASTyLinker is expected to grow proportionally to the runtime of BLINK \cite{wu2020scalable} for increasing sizes of datasets. If runtime is an important factor, one might consider skipping entity reranking as NASTyLinker still produces reasonable results when relying on bi-encoder affinities only.
\subsection{Linking Entities in Wikipedia Listings}
\label{sec:linking-entities-in-wikipedia-listings}
As the average mention context length in the LISTING dataset is lower than the one in NILK, fine-tuning the EL models took only a total of 8 hours. We find that most of the hyperparameters chosen for NILK are a reasonable choice for this dataset as well. For entity reranking, however, the approaches produce better results when the thresholds $\tau_m$ and $\tau_a$ are slightly increased to 0.9 and 0.85.
\subsubsection{Results on Test Partition}
Linking results for $\mathcal{D}^L_{test}$ are provided in Table~\ref{tab:listing-results}. As we rely on PCA for the labelling of NIL mentions, we only know whether a mention is a NIL mention without knowing which NIL mentions refer to the same entity. Hence, we can only compute results for known entities and for overall predictions. For the latter, we simply assume that any prediction made for a NIL mention is incorrect. With this assumption, we are obviously not able to produce realistic performance estimates, but we are able to see the impact of being NIL-aware (and hence, make no prediction for NIL mentions) on the overall linking performance.

Due to their majority mechanism, Majority Clustering identifies known entities with very high precision, but at the cost of a reduced recall. The scores of Bottom-Up Clustering and NASTyLinker are comparable when considering known entities, but diverge w.r.t. the micro average. In the entity reranking scenario, NASTyLinker achieves the overall best micro F1-score with 86.7\%. This, however, has to be taken with a grain of salt as we do not know how many of the heuristically labelled NIL mentions are actually referring to NIL entities and how many refer to known entities.

{\renewcommand{\arraystretch}{1.05}%
    \begin{table}
      \centering
      \caption{Results for the test partition $\mathcal{D}^{L}_{test}$ of the LISTING dataset. No results for \textit{NIL} are given because the real NIL entities $\mathcal{E}^n$ are not available for this dataset. For the micro average, we label every prediction made for a mention linked to an entity in $\mathcal{E}^n$ as incorrect.}
      \label{tab:listing-results}
      \begin{tabular}{|C{2cm}|C{2.5cm}|C{0.8cm}|C{0.8cm}|C{0.8cm}|C{0.8cm}|C{0.8cm}|C{0.8cm}|}
        \hline
        \multicolumn{2}{|c|}{\textbf{Approach}} & \multicolumn{3}{c|}{\textbf{Known}} & \multicolumn{3}{c|}{\textbf{Micro}} \\
        \multicolumn{2}{|c|}{} & P & R & F1 & P & R & F1 \\
        \hline \hline
        \multirow{3}{*}{\shortstack{\textbf{No}\\\textbf{Clustering}}} & Exact Match & 91.4 & 73.5 & 81.5 & 81.1 & 73.5 & 77.1 \\
         & Bi-Encoder & 88.6 & 88.6 & 88.6 & 62.6 & 88.6 & 73.4 \\
         & Cross-Encoder & 93.7 & \textbf{93.8} & \textbf{93.8} & 66.2 & \textbf{93.8} & 77.6 \\
        \hline
        \multirow{3}{*}{\shortstack{\textbf{Clustering}\\\textbf{\& No}\\\textbf{Reranking}}}
         & Bottom-Up & 89.7 & 84.9 & 87.2 & 63.9 & 84.9 & 72.9 \\
         & Majority & 95.2 & 67.9 & 79.2 & 78.1 & 67.9 & 72.6 \\
         & NASTyLinker & 90.6 & 78.5 & 84.1 & 70.7 & 78.5 & 74.4 \\
        \hline
        \multirow{3}{*}{\shortstack{\textbf{Clustering}\\\textbf{\& Entity}\\\textbf{Reranking}}}
         & Bottom-Up & 94.2 & 90.8 & 92.5 & 75.3 & 90.8 & 82.3 \\
         & Majority & \textbf{98.8} & 76.2 & 86.0 & \textbf{93.4} & 76.2 & 83.9 \\
         & NASTyLinker & 97.0 & 87.0 & 91.8 & 88.5 & 87.0 & \textbf{87.7} \\
         \hline
    \end{tabular}
    \end{table}
}
\subsubsection{Knowledge Graph Population Statistics}
The partition $\mathcal{D}^L_{pred}$ of the LISTING dataset contains only mentions for which we don't know whether they link to a known or to a NIL entity. To make predictions for these mentions, we run the NASTyLinker approach on the whole LISTING corpus, i.e. on a total of 38 million mentions, as we need representations of all known entities for the clustering step. These mentions were extracted from 2.9 million listings on 1.4 million Wikipedia pages. As reference KB, we use the knowledge graph CaLiGraph which is based on Wikipedia and hence contains entities for all 5.8 million Wikipedia articles.

The total runtime was 62 hours, with 14 hours for the bi-encoder, 47 hours for the cross-encoder, and 45 minutes for the clustering. We find 13.4 million mentions (i.e., 70\%) to be NIL mentions which refer to 7.6 million NIL entities. The remaining 5.2 million mentions refer to 1.4 million entities that exist in CaLiGraph already. By integrating the discovered NIL entities into CaLiGraph, we would increase its entity count by 130\%. Further, the discovered mentions for known entities can be used to enrich the representations of the entities in the knowledge graph through various knowledge graph completion methods \cite{heist2021information}.
\subsubsection{Qualitative Analysis}
To evaluate the actual linking performance on the set of unlabeled mentions $\mathcal{D}^L_{pred}$, we conducted a manual inspection of the results. We randomly picked 100 mentions and 100 clusters\footnote{The sampling of clusters was stratified w.r.t. cluster size.} and identified, if incorrect, the type of error.\footnote{We evaluated the linking and clustering decision w.r.t. the top-4 mention and entity candidates produced by the bi-encoder. Although recall@4 for the bi-encoder is 97\%, some relevant candidates might have been missed.} The results of this evaluation are given in Table~\ref{tab:manual-evaluation}. Overall, we find the outcome to agree with the results of NASTyLinker on $\mathcal{D}^L_{test}$. Hence, the approach produces highly accurate results, which we observed even for difficult cases. For example, the approach correctly created NIL entity clusters for the mention \textit{North Course} referring to a racing horse (in pages \textit{Appleton Stakes} and \textit{Oceanport Stakes}), a golf course in Ontario, CA (in page \textit{Tournament Players Club}), and a golf course in Florida, US (in page \textit{Pete Dye}).

{\renewcommand{\arraystretch}{1.05}%
    \begin{table}[t]
      \centering
      \caption{Results of the manual evaluation of 100 clusters and 100 mentions. Columns group the results by actual entity type (known, NIL, overall), rows group by prediction outcome. Accuracy values may deviate by $\pm$9.6\% for mentions and by $\pm$7.0\% for clusters (95\% confidence).}
      \label{tab:manual-evaluation}
      \begin{tabular}{|C{6.2cm}|C{0.8cm}|C{0.8cm}|C{0.8cm}|C{0.8cm}|C{0.8cm}|C{0.8cm}|}
        \hline
        \textbf{Prediction} & \multicolumn{3}{c|}{\textbf{Mentions}} & \multicolumn{3}{c|}{\textbf{Clusters}} \\
        \textbf{} & $\mathcal{E}^k$ & $\mathcal{E}^n$ & $\mathcal{E}$ & $\mathcal{E}^k$ & $\mathcal{E}^n$ & $\mathcal{E}$ \\
        \hline \hline
        Correct & 20 & 64 & 84 & 8 & 71 & 79 \\
        Incorrectly linked to NIL entity & 3 & --- & 3 & 1 & --- & 1 \\
        Incorrectly linked to known entity & --- & 7 & 7 & --- & 3 & 3 \\
        Not all mentions of entity in cluster & --- & --- & --- & 8 & 0 & 8 \\
        Mentions from multiple entities in cluster & --- & --- & --- & 1 & 4 & 5 \\
        \hline
        \hline
        Ignored (mention extracted incorrectly) & --- & --- & 6 & --- & --- & 4 \\
        Total Count & 23 & 71 & 94 & 18 & 78 & 96 \\
        Accuracy (\%) & 87.0 & 90.1 & 89.4 & 44.4 & 91.0 & 82.3 \\
        \hline
    \end{tabular}
    \end{table}
}

While the linking performance is quite consistent for mentions, the correctness of clusters for known entities is significantly lower than for NIL entities.\footnote{For the evaluation to be significant, we treat all clusters referring to the same known entity as a single cluster.} This drop in performance is not due to NASTyLinker being incapable of linking to known entities correctly (as the accuracy of 87\% on mention-level shows). Instead, it can rather be attributed to the fact that clusters of known entities contain 3.8 mentions on average, while clusters of NIL entities contain 1.7 mentions on average. Hence, the likelihood of missing at least one mention is a lot higher, which is also the main error for known clusters.

Compared to the results on NILK, the linking accuracy for NIL mentions is much higher. We explain this with the different kinds of NIL entities contained in the two datasets. While an average NIL entity is mentioned 4.6 times in NILK, our results indicate that this number is approximately 1.7 for the LISTING dataset. The latter dataset may hence contain a lot of easy-to-link mentions by assigning them their own cluster.
\section{Conclusion and Outlook}
With NASTyLinker, we introduce a NIL-aware EL approach that is capable of making high-quality predictions for known and for NIL entities. In the practical setting of EL in Wikipedia listings, we show that our approach can be used to populate a knowledge graph with a large number of additional entities as well as to enrich representations of existing entities.

Although the results look promising at a first glance, there is still a lot to improve as even small errors can multiply in downstream applications. For future work, we plan to concentrate on establishing a full end-to-end pipeline that includes the detection of mention, as recent works demonstrate how this can substantially reduce runtime without a decrease in performance \cite{ayoola2022refined,kassner2022edin}. This will also open the path to a training procedure that considers NIL entities already during the creation of embeddings. Additionally, we will explore how the dependencies between items in listings can be exploited to further improve predictions.
\bibliographystyle{splncs04}
\bibliography{references}
\end{document}